\documentclass{article}

\usepackage{graphics}
\usepackage{graphicx}
\usepackage{multicol}
\usepackage{multirow}
\usepackage{soul}
\usepackage{changes}
\usepackage{url}
\usepackage{amssymb}

\title{Symlink: A New Dataset \\for Scientific Symbol-Description Linking}

\author{\textbf{$^1$Viet Dac Lai, $^1$Amir Pouran Ben Veyseh, }\\\textbf{$^2$Franck Dernoncourt, $^1$Thien Huu Nguyen}\\
$^1$Dept. of Computer and Information Science\\
University of Oregon\\ 
$^2$Adobe Research\\
\{vietl,apouran,thien\}@cs.uoregon.edu\\franck.dernoncourt@adobe.com
 }

\date{}

\begin{document}

\maketitle

\begin{abstract}
Mathematical symbols and descriptions appear in various forms across document section boundaries without explicit markup. Linking mathematical symbols and their descriptions has been conducted for a decade; however, the current state of research is distant toward real application. In this paper, we present a new large-scale dataset that emphasizes extracting symbols and descriptions in scientific documents. Symlink annotates scientific papers of 5 different domains (i.e., computer science, biology, physics, mathematics, and economics). Our experiments on Symlink demonstrate the challenges of the symbol-description linking task for existing models and call for further research effort in this area. We will publicly release Symlink to facilitate future research. 
\end{abstract}

\newcommand{\mytitle}[2]{\multicolumn{#1}{|c|}{\textbf{#2}}}

\section{Introduction}

The exponential growth of published articles exceeds any individual ability to keep track of the development of the field. Hence, automatic reading comprehension of scientific documents has attracted the attention of researchers across various domains such as Drug Discovery, Knowledge Base Construction, and Natural Language Processing. A pivotal angle of understanding scientific literature is understanding their terminologies and the corresponding mathematical formulae because the terminologies and formulae offer an explicit, precise interface to present the relation between scientific concepts \cite{schubotz2018improving}. As such, in order to understand the scientific concepts, a reading comprehension machine needs to (i) identify the scientific concept descriptions and their formulae, (ii) segment them into primitive terms and symbols, and (iii) link the associated terms and the corresponding symbols.

Working with mathematical formulae is arduous due to two fundamental reasons. First, common text encodings such as ASCII and Unicode do not fully support typing mathematical symbols.  As a result, it is impossible to write complex mathematical formulae in plain text using either ASCII or Unicode. As such, a higher level encoding/typesetting is essential to store the content of scientific documents, such as LaTex. Second, most of the scientific documents are stored in one of two forms: photos or Portable Document Format (PDF). Scientific documents that were published prior to the graphical computer era are printed and now scanned as photos. Since the development of graphical computers, scientific documents have been composed and exported to PDF format. Unfortunately, analyzing textual information in photo images or PDF files is extremely difficult, and most of the NLP tools are not developed to handle this type of data. As such, to facilitate the understanding of scientific literature, the documents should be stored using a universal easy-to-process text-like encoding. In this paper, we propose to use LaTeX as the base typesetting to enable the whole analyzing pipeline of scientific documents.

LaTeX is a high-quality typesetting system that is based on the ``\textbf{what you see is what you mean}'' concept, meaning the writers just need to focus on the content of the documents; the computer will take care of the formatting. LaTeX is widely used in producing scientific documents, books, and many other publishing products. LaTeX is a powerful typesetting that it not only produces beautiful documents, but also allows writers to quickly compose complicated document objects such as mathematical symbols and formulae, tables, and graphs. Once a LaTeX document is completed, a LaTeX compiler produces a PDF file with magnificent formatting and those complex document objects. Thanks to recent advances in text processing and image recognition, a LaTeX document can be restored at a high quality from either a photo or a PDF file \cite{deng2017image}. 

\begin{figure*}
    \centering
    \includegraphics[width=\textwidth]{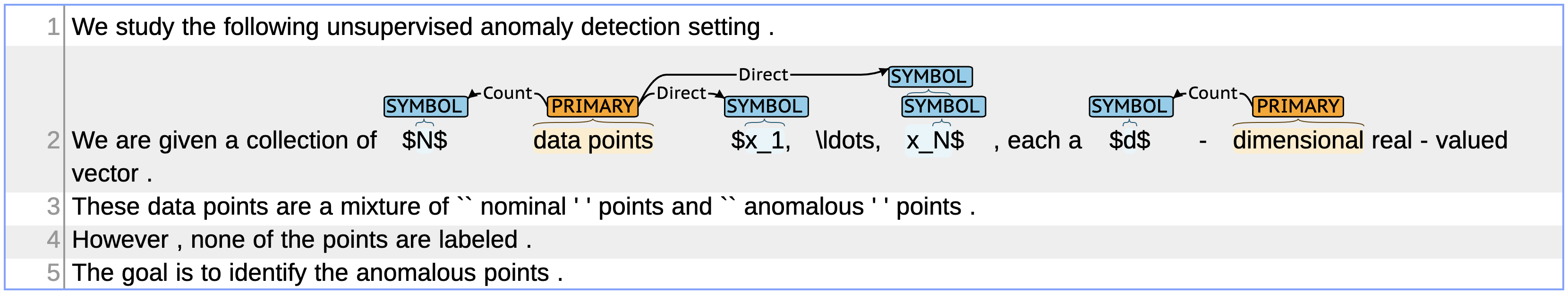}
    \caption{An example of the LaTex document of a scientific article with annotated relations.}
    \label{fig:example}
\end{figure*}

In this paper, we focus on extracting mathematical symbols and their corresponding textual descriptions from the LaTeX documents. In particular, given a paragraph as shown in Figure \ref{fig:example}, multiple symbols are described within the span of the second sentence, such as the number of data points $N$, individual data points $x_1, and x_N$, and vector dimension $d$. Fortunately, the descriptions of these symbols are expressed within the scope of the mentioned sentence. As such, the goal of this work is to link all the symbols and their corresponding descriptions within the context.

Many studies have attempted the problem in various ways. Mathematical symbols can be recognized and linked to Wikipedia pages \cite{nghiem2010mining,kristianto2016entity}. Symbols and descriptions in close proximity can be linked \cite{kristianto2014extracting,alexeeva2020mathalign} because the descriptions are dedicated to the symbols and the context presented in the document. Prior studies attempted to link the description to expressions at formula-level \cite{nghiem2010mining,kristianto2014extracting, kristianto2016entity}. However, it is essential to understand mathematical expression at a more primitive level, e.g. symbol level. Our work focuses on this fine-grained level annotation for mathematical symbol/description extraction and linking to drive future research toward a better understanding of scientific documents. Moreover, since the prior methods are based on pattern matching and rule-based algorithms, a few small-scale datasets are created for evaluation purpose only on English literature. As the result, it is impossible to train and evaluate advanced deep learning-based models on these small datasets. This paper presents a large-scale dataset for Symbol Description Linking task for English scientific literature.

The contribution of this paper is three-fold:

\begin{itemize}
    \item We propose to use LaTex as the intermediate format for scientific documents.
    \item We create the first large-scale dataset for Symbol Description Linking problem. 
    \item We evaluate the state-of-the-art joint entity-relation extraction models on the new dataset.
\end{itemize}

\section{Data Annotation}

We present the annotation process to create a large-scale dataset for Symbol-Description linking in scientific documents.

\textbf{Data source}: We obtain the documents from \url{arXiv.org}, a repository for preprint scientific articles due to the broad coverage of subjects in scientific articles published in ArXiv. In particular, ArXiv offers articles in physics, mathematics, quantitative biology, computer science, quantitative finance, statistics, electrical engineering, and economics. As such, our obtained papers contain a large number of mathematical symbols and equations, allowing a higher yield of extracted symbol-description relations. Among these subjects, we choose five subjects of mathematics, physics, biology, economics, and computer science for annotation.

\textbf{Data preparation}: ArXiv open-sources the LaTeX version of their articles, when available. In order to make our Symlink dataset open-access to the whole community, we crawl the metadata of these articles and only selected articles under the CC BY license. Once obtained the LaTeX project, we extract all the paragraphs from the \textbf{.tex} files. We filter out all short paragraphs with less than 50 words and paragraphs without symbols. Since a formula can be composed in multiple ways such as inline formulae (between $\$\;\;\$$), displayed formulae (between $\$\$\;\;\$\$$), or using commands e.g. \textit{array}, to keep the original TeX format of the formulae, all of these math objects are masked before tokenization. Then, we use the tokenizer from SciBERT \cite{beltagy-etal-2019-scibert} to tokenize the text. The original math object is then restored. As we observe that many papers have nested math objects, we delete all the nested objects, hence, having non-nested LaTeX data. This is helpful as it makes the LaTeX documents more similar to the ones generated by the PDF-to-LaTeX tools, which does not contain nested objects.

\textbf{Annotation taxonomy}: To prepare for the annotation, we design a taxonomy with 3 general entity types and four relation types. In particular, mathematical symbols are annotated under the tag \textbf{SYMBOL}, whereas descriptions are tagged under two labels \textbf{PRIMARY}, for single standalone definitions, and \textbf{ORDERED}, for the description of multiple terms, whose mentions are not separated without creating non-contiguous mentions. Due to the massive amount of combinations of descriptions and complex math expressions, we only tag an entity if and only if there is a second entity that pairs with the first entity to form a relation. For relation, we are particularly interested in two main types of relations: \textbf{DIRECT}, linking a symbol with its definition, and \textbf{COUNT}, linking a description of a concept with a symbol that is the number of instances of the concept. Due to the sheer number of repetitions and coreferences of both descriptions and symbols, we also annotate \textbf{COREF-SYMBOL} relation, linking corefered symbols, and \textbf{COREF-DESCRIPTION} relation, linking corefered descriptions. Detailed annotation guildlines with examples are presented in Appendix \ref{app:guidelines}.

\textbf{Annotation}: We recruited 10 annotators from the crowdsourcing platform \url{upwork.com} to annotate scientific papers in the five mentioned domains (each subject was annotated by two annotators). Since Upwork allows freelancers to submit a resume for each job application, we explicitly select annotators that have demonstrated experience in reading and writing scientific papers in their corresponding domains (e.g., holding an M.S. or Ph.D. degree). Detailed annotation guidelines with many examples and explanations are provided to train the annotators. Overall, we annotated 101 papers, accounting for 5,719 sentences, and 330K tokens. Our annotators for each domain co-annotate the documents in their domain and achieve Cohen's Kappa scores of (averaged) 0.79. This inter-agreement score thus indicates substantial agreements between our annotators. Eventually, the annotators are engaged in discussions to resolve any conflict to produce a final consolidated version of our Symlink dataset. 
\begin{table}[t]
\centering
\caption{Statistics and label distribution of the Symlink dataset. $^*$The texts are tokenized by SciBERT tokenizer.}
\begin{tabular}{|l|r|r|r|r|}
    \hline
     & \mytitle{1}{Train} & \mytitle{1}{Dev} & \mytitle{1}{Test} & \mytitle{1}{Total}\\ \hline 
     \mytitle{5}{Statistics} \\
     \hline 
    \#Documents  & 91 & 6 & 5 & 102 \\
    \#Paragraphs & 2,097 & 201 & 213 & 2,511\\
    \#Sentences  & 4,794 & 412 & 513 & 5,719\\
    \#Tokens$^*$ & 283,491 & 23,036 & 28,395 & 334,922\\
    \hline
    \mytitle{5}{Entity types} \\
    \hline
    \#SYMBOL & 11,404 & 1,199 & 1,231 & 13,834\\
    \#PRIMARY & 7,095 & 626 & 762 & 8,483 \\
    \#ORDERED & 13 & 3 & 1 & 17 \\
    \hline
    \mytitle{5}{Relation types} \\
    \hline
    \#Direct            & 7,186 & 703    & 673   & 8,562\\
    \#Coref-Symbol      & 2,658 & 416    & 302   & 3,376 \\
    \#Coref-Description & 243   & 55     & 51    & 349   \\
    \#Count             & 1,010 & 17     & 152   & 1,179\\
    \hline
\end{tabular}
\label{tab:statistics}
\end{table}

\section{Dataset Analysis}

Table \ref{tab:statistics} presents the statistics for the dataset including the number of articles, distribution of entities, and distribution of the relations. Overall, our dataset offers more than 22K entities, 10K pairs of description and symbol, which is one order of magnitude larger than existing datasets for the similar task.

\begin{figure}[!h]
    \centering
    \includegraphics[width=\linewidth]{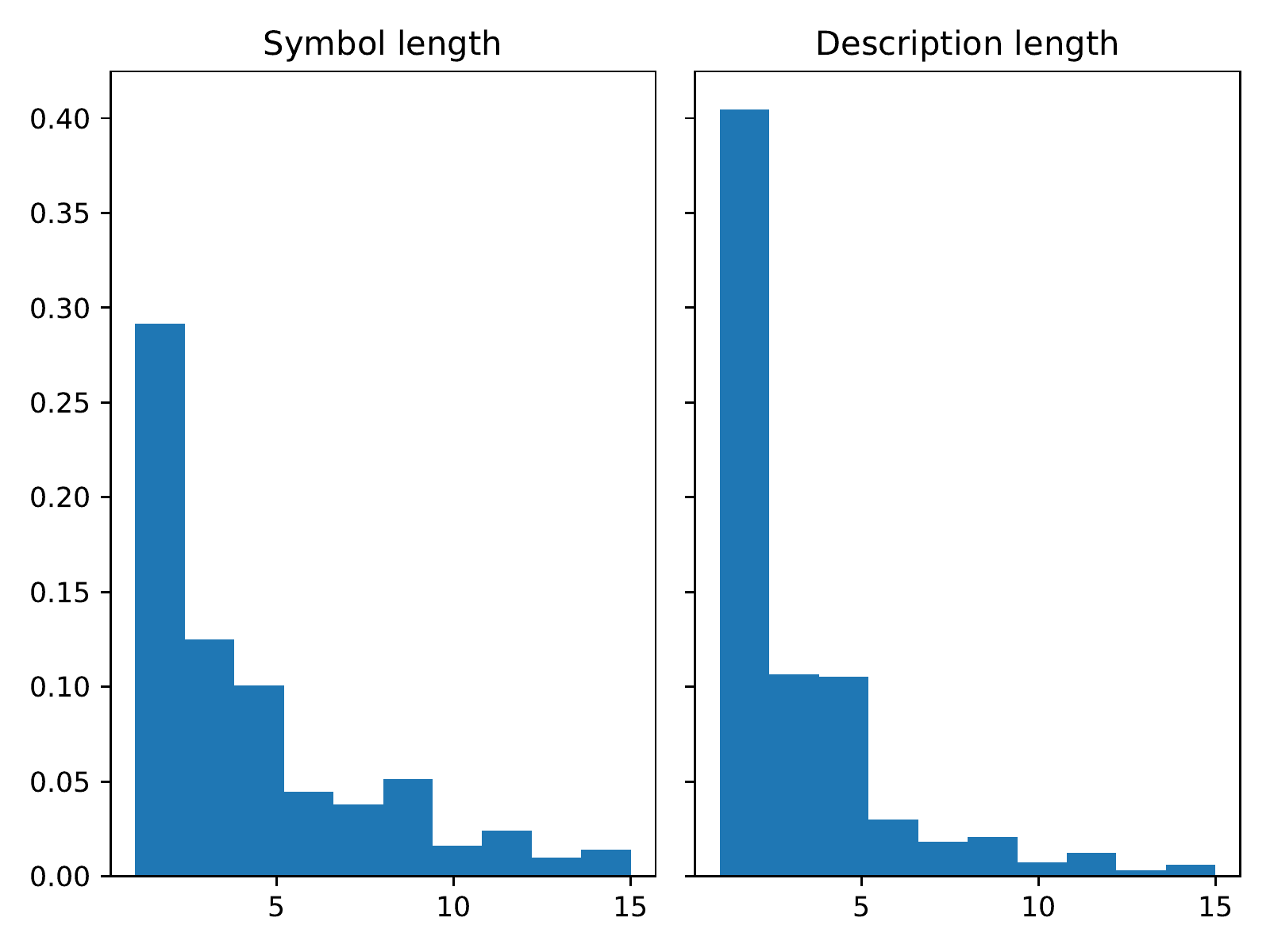}
    \caption{Length of symbols and descriptions in Symlink}
    \label{fig:span-length}
\end{figure}

Figure \ref{fig:span-length} presents the distribution of the span lengths of both symbols and descriptions of up to 15 tokens. As can be seen from the figure, the majority of entities has a length of 1-3 tokens. However, overall, the span lengths of both symbols and descriptions vary significantly from 1 up to 47 tokens (note that Figure \ref{fig:span-length} only illustrates the spans with up to 15 tokens). 
This demonstrates a key challenge of the Symbol-Description Linking task in this paper where symbols and descriptions with long spans might introduce confusion for extraction models.

\begin{figure}[!h]
    \centering
    \includegraphics[width=\linewidth]{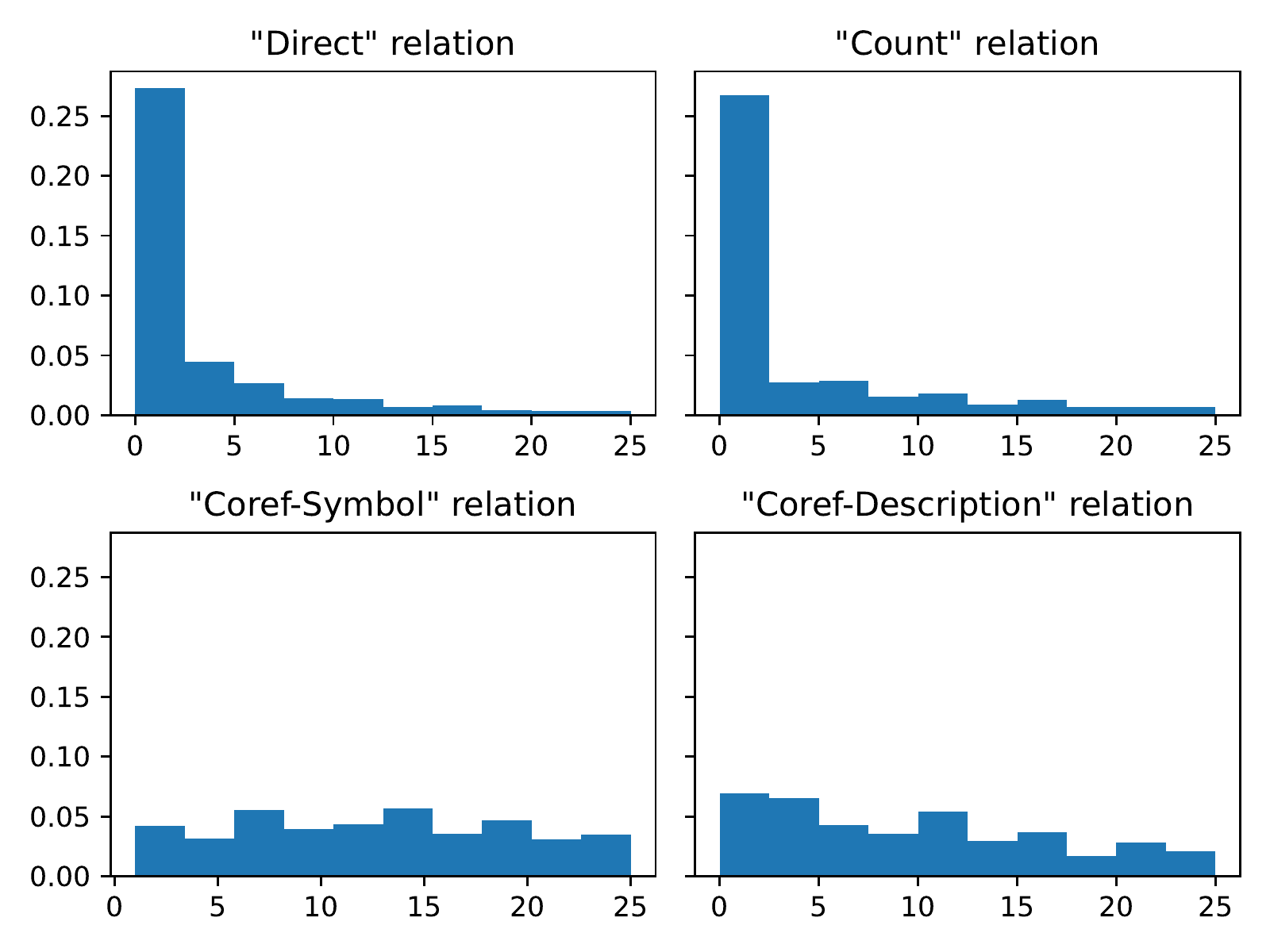}
    \caption{Distribution of distances between entities in Symlink by relation type.}
    \label{fig:distance}
\end{figure}

To further understand the dataset, we present the distances between the entities and relations annotated in Symlink by different relation types in Figure  \ref{fig:distance}. The distributions can be group in two categories. The first category involves the symbol-description relations while the second group involves the coreference relations. The distributions of symbol-description relations have long tails, indicating that symbols and descriptions tend to appears in a close proximity. On the other hand, the distributions of coreference relations are quite flat, suggesting that coreference relation relationship appears in both short and long distances.

\section{Experiments}

To assess the complexity of the Symbol Description Linking task in Symlink, we evaluate the performance of the state-of-the-art deep learning models for ED. In particular, the following models are tested:

\begin{itemize}

    \item \textbf{OneIE} \cite{lin-etal-2020-joint}: OneIE is one of the state-of-the-art models in information extraction (IE) that is designed to jointly perform four tasks: entity mention detection, relation extraction, event detection, and argument role labeling. We adapt OneIE to our work using two tasks, i.e., entity mention detection and relation extraction. The model first detects the spans of entity mentions in the text. The detected spans are then paired to jointly predict entity types and relations. 
    \item \textbf{FourIE} \cite{nguyen-etal-2021-cross}: FourIE is another state-of-the-art models for IE to simultaneously solves the four IE tasks in a single model. In addition, FourIE captures the inter-dependencies between prediction instances of different tasks by letting their representations interact with each other, hence, enhancing the predictions of the model. Similar to OneIE, we retain the two modules that perform entity mention detection and relation extraction, thus turning the model into a joint model for entity-relation extraction. This model is the best model on the ACE++ dataset \cite{Walker:05}.

    \item \textbf{Jerex} \cite{eberts-ulges-2021-end} is a joint model for entity-level relation extraction from documents. Jerex introduces a multi-task approach that builds upon coreference resolution and gathers relevant signals via multi-instance learning with multi-level representations, combining global entity and local mention information. The model achieves state-of-the-art relation extraction results on the DocRED dataset \cite{yao-etal-2019-docred}.

\end{itemize}

\begin{table}[t]
    \centering
    \caption{Performance of the state-of-the-art models on Symlink dataset with references of performance on DocRED and ACE++ datasets. The numbers in the right most column show the performances of Jerex on DocRED dataset and OneIE and FourIE on ACE++ dataset. P, R, F stand for precision, recall, f1-score, respectively.}
    \begin{tabular}{|c|c|ccc|ccc|c|}
        \hline
        \multirow{2}{*}{\textbf{Model}} & \multirow{2}{*}{\textbf{Task}} & \mytitle{3}{Dev} & \mytitle{3}{Test} & \multirow{2}{*}{\textbf{F}}\\
        \cline{3-8}
        & & P & R & F & P & R & F & \\
        \hline
        \multirow{2}{*}{Jerex} & NER & \textbf{82.0} & 60.8 & 69.8 & \textbf{60.9} & 35.7 & 45.0 & 80.1\\
        & RE & \textbf{72.3} & 44.5 & 55.1 & 46.9 & 20.6 & 28.7 & 40.4\\ 
        \hline
        \multirow{2}{*}{OneIE} & NER &76.9 & \textbf{69.5} & \textbf{73.0} & 58.8 & 51.3 & \textbf{54.8} & 89.6\\
        & RE &61.4 & \textbf{54.2} & \textbf{57.5} & 34.2 & \textbf{31.1} & \textbf{32.6} & 58.6\\ 
        \hline
        \multirow{2}{*}{FourIE} & NER &75.3 & 68.4 & 71.7 & 55.1 & \textbf{52.2} & 53.6 & 91.1\\
        & RE &59.7 & 50.5 & 54.7 & 28.7 & 26.6 & 27.6 & 63.6\\
        \hline
    \end{tabular}
    \label{tab:result}
\end{table}

\textbf{Hyper-parameters}: As the texts in this dataset are obtained from scientific documents, we use the \textit{allenai/scibert\_scivocab\_cased} version of SciBERT in all the considered models \cite{beltagy-etal-2019-scibert}. The Stanza toolkit \cite{qi-etal-2020-stanza}\footnote{https://stanfordnlp.github.io/stanza/} is employed for sentence splitting and dependency parsing. The hyper-parameters of the trained models are tuned over the development set based the suggested values from the original papers.


\textbf{Results}: Table \ref{tab:result} reports the performance of the
models on the development and test sets of Symlink over two tasks: Entity Mention Detection (Entity) and Relation Extraction (Relation). Among the examined models, OneIE offers the best performance with F1 scores of 54.8\% for entity recognition and 32.6\% for relation extraction. For comparison, we also include the reported performances of these models on the DocRED and ACE++ datasets \cite{nguyen-etal-2021-cross,yao-etal-2019-docred}. As can be seen, the performance of the current information extraction models on Symlink is significantly and substantially worse than those on the DocRED and ACE++ datasets. For the entity recognition task, the F1-score difference between Symlink and DocRED for Jerex is 35\%, and those between Symlink and ACE++ for OneIE is 35\%. Due to the low performance of entity recognition, the relation extraction performance hurts sharply by up to 36\% between Symlink and ACE++ of the FourIE model. This suggests that the Symbol-Description Linking task in Symlink is a challenging task and more research effort is necessary to boost the performance of the models for this topic.


\section{Related Work}

Early studies for scientific literature link formulae to Wikipedia page \cite{nghiem2010mining,kristianto2016entity}. Even though this can provide additional information regarding the mathematical expression, a reader might find it harder to understand the Wikipedia page as it is presented in many unrelated forms. Linking to the description in the same document is more practical \cite{kristianto2014extracting,alexeeva2020mathalign} as the descriptions are dedicated to the symbols and the context presented in the document. 

Previous studies on symbol-description extraction rely on pattern matching \cite{yokoi2011contextual,nghiem2010mining} and rule-based algorithms \cite{alexeeva2020mathalign}. These methods might work for observed patterns with an assumption of close proximity between symbol and description. They may fail to capture distant symbol-description pairs and symbols in very complex structures such as algorithms in computer science literature. 

Most of the previous studies have attempted to extract and link at formula level \cite{nghiem2010mining,kristianto2014extracting, kristianto2016entity}. In reality, understanding mathematical formulae requires details of atomic symbols e.g. superscript, subscript, function arguments. We believe that address the problem at this fine-grain level is crucial to drive future research toward a better understanding of the complex symbol-description extraction task.

There have been some datasets created Symbol-Description Linking \cite{yokoi2011contextual}. However, one of them is created for publications written in Japanese \cite{yokoi2011contextual}, making it nearly impossible to transfer to English literature. Moreover, two other datasets \cite{schubotz2016semantification,alexeeva2020mathalign} only annotate a small-scale golden data for evaluation purpose. As the result, no training data is available for training deep neural network models. In this paper, we provide a large-scale dataset for English literature that we believe will provide enough supervision for the promising deep neural network-based models.

\section{Conclusion}

We present Symlink, the first large-scale dataset for Symbol-Description Linking on scientific documents. Our experiments demonstrate the poor performance of current information extraction and relation extraction models on the Symlink dataset. This dataset also standardizes the Symbol-Description Linking task, hence facilitating the future research in this topic. In the future, we plan to enlarge the dataset to include more entity and relation types for scientific documents.

\bibliographystyle{acm}
\bibliography{anthology,ref}

\begin{thebibliography}{10}

\bibitem{alexeeva2020mathalign}
{\sc Alexeeva, M., Sharp, R., Valenzuela-Esc{\'a}rcega, M.~A., Kadowaki, J.,
  Pyarelal, A., and Morrison, C.}
\newblock {M}ath{A}lign: Linking formula identifiers to their contextual
  natural language descriptions.
\newblock In {\em Proceedings of the 12th Language Resources and Evaluation
  Conference\/} (Marseille, France, May 2020), European Language Resources
  Association, pp.~2204--2212.

\bibitem{beltagy-etal-2019-scibert}
{\sc Beltagy, I., Lo, K., and Cohan, A.}
\newblock {S}ci{BERT}: A pretrained language model for scientific text.
\newblock In {\em Proceedings of the 2019 Conference on Empirical Methods in
  Natural Language Processing and the 9th International Joint Conference on
  Natural Language Processing (EMNLP-IJCNLP)\/} (Hong Kong, China, Nov. 2019),
  Association for Computational Linguistics, pp.~3615--3620.

\bibitem{deng2017image}
{\sc Deng, Y., Kanervisto, A., Ling, J., and Rush, A.~M.}
\newblock Image-to-markup generation with coarse-to-fine attention.
\newblock In {\em International Conference on Machine Learning\/} (2017), PMLR,
  pp.~980--989.

\bibitem{eberts-ulges-2021-end}
{\sc Eberts, M., and Ulges, A.}
\newblock An end-to-end model for entity-level relation extraction using
  multi-instance learning.
\newblock In {\em Proceedings of the 16th Conference of the European Chapter of
  the Association for Computational Linguistics: Main Volume\/} (Online, Apr.
  2021), Association for Computational Linguistics, pp.~3650--3660.

\bibitem{kristianto2014extracting}
{\sc Kristianto, G.~Y., Aizawa, A., et~al.}
\newblock Extracting textual descriptions of mathematical expressions in
  scientific papers.
\newblock {\em D-Lib Magazine 20}, 11 (2014), 9.

\bibitem{kristianto2016entity}
{\sc Kristianto, G.~Y., Topi{\'c}, G., and Aizawa, A.}
\newblock Entity linking for mathematical expressions in scientific documents.
\newblock In {\em International Conference on Asian Digital Libraries\/}
  (2016), Springer, pp.~144--149.

\bibitem{lin-etal-2020-joint}
{\sc Lin, Y., Ji, H., Huang, F., and Wu, L.}
\newblock A joint neural model for information extraction with global features.
\newblock In {\em Proceedings of the 58th Annual Meeting of the Association for
  Computational Linguistics\/} (Online, July 2020), Association for
  Computational Linguistics, pp.~7999--8009.

\bibitem{nghiem2010mining}
{\sc Nghiem~Quoc, M., Yokoi, K., Matsubayashi, Y., and Aizawa, A.}
\newblock Mining coreference relations between formulas and text using
  {W}ikipedia.
\newblock In {\em Proceedings of the Second Workshop on {NLP} Challenges in the
  Information Explosion Era ({NLPIX} 2010)\/} (Beijing, China, Aug. 2010),
  Coling 2010 Organizing Committee, pp.~69--74.

\bibitem{nguyen-etal-2021-cross}
{\sc Nguyen, M.~V., Lai, V., and Nguyen, T.~H.}
\newblock Cross-task instance representation interactions and label
  dependencies for joint information extraction with graph convolutional
  networks.
\newblock In {\em Proceedings of the 2021 Conference of the North American
  Chapter of the Association for Computational Linguistics: Human Language
  Technologies\/} (Online, June 2021), Association for Computational
  Linguistics, pp.~27--38.

\bibitem{qi-etal-2020-stanza}
{\sc Qi, P., Zhang, Y., Zhang, Y., Bolton, J., and Manning, C.~D.}
\newblock {S}tanza: A python natural language processing toolkit for many human
  languages.
\newblock In {\em Proceedings of the 58th Annual Meeting of the Association for
  Computational Linguistics: System Demonstrations\/} (Online, July 2020),
  Association for Computational Linguistics, pp.~101--108.

\bibitem{schubotz2018improving}
{\sc Schubotz, M., Greiner-Petter, A., Scharpf, P., Meuschke, N., Cohl, H.~S.,
  and Gipp, B.}
\newblock Improving the representation and conversion of mathematical formulae
  by considering their textual context.
\newblock In {\em Proceedings of the 18th ACM/IEEE on Joint Conference on
  Digital Libraries\/} (2018), pp.~233--242.

\bibitem{schubotz2016semantification}
{\sc Schubotz, M., Grigorev, A., Leich, M., Cohl, H.~S., Meuschke, N., Gipp,
  B., Youssef, A.~S., and Markl, V.}
\newblock Semantification of identifiers in mathematics for better math
  information retrieval.
\newblock In {\em Proceedings of the 39th International ACM SIGIR conference on
  Research and Development in Information Retrieval\/} (2016), pp.~135--144.

\bibitem{Walker:05}
{\sc Walker, C., Strassel, S., Medero, J., and Maeda, K.}
\newblock Ace 2005 multilingual training corpus.
\newblock In {\em Technical report, Linguistic Data Consortium\/} (2006).

\bibitem{yao-etal-2019-docred}
{\sc Yao, Y., Ye, D., Li, P., Han, X., Lin, Y., Liu, Z., Liu, Z., Huang, L.,
  Zhou, J., and Sun, M.}
\newblock {D}oc{RED}: A large-scale document-level relation extraction dataset.
\newblock In {\em Proceedings of the 57th Annual Meeting of the Association for
  Computational Linguistics\/} (Florence, Italy, July 2019), Association for
  Computational Linguistics, pp.~764--777.

\bibitem{yokoi2011contextual}
{\sc Yokoi, K., Nghiem, M.-Q., Matsubayashi, Y., and Aizawa, A.}
\newblock Contextual analysis of mathematical expressions for advanced
  mathematical search.
\newblock {\em Polibits}, 43 (2011), 81--86.

\end{thebibliography}

\appendix
\clearpage

\section{Annotation guidelines}
\label{app:guidelines}

This section summarizes some rules that we use to make our annotations more consistent. Examples are provided in Table \ref{tab:examples}.

\textbf{Description tagging}: A description is usually a noun or a noun phrase that expresses a concept. These are the overall rules for entity annotations:
\begin{itemize}
    \item We only tag a description if the corresponding symbol presents in the text.
    \item A description usually is a noun or a noun phrase. Sometimes, a verb, an adverb, or an adjective describes an operation, it is also considered a description.
    \item Descriptions should be short but it must cover the elements in the corresponding symbol, esp. in case of complex symbols, such as superscript, subscript, arguments, and limits.
\end{itemize}

\newcommand{\hla}[1]{\colorbox{orange}{#1}}
\newcommand{\hlb}[1]{\colorbox{yellow}{#1}}
\newcommand{\hlc}[1]{\colorbox{cyan}{#1}}
\newcommand{\rota}[2]{\parbox[t]{2mm}{\multirow{#1}{*}{\rotatebox[origin=c]{90}{\textbf{#2}}}}}

\begin{table}[!h]
    \centering
    \caption{Tag schema for description recognition. The tag name and the entity mentions are highlighted accordingly.}
    \begin{tabular}{|l|l|p{8cm}|}
        \hline
       \rota{3}{Entity} & \hlc{Symbol} &  \hla{This set of class} is divided into \hlb{base and novel classes} \$\hlc{C\_b}\$ and \$\hlc{C\_n}\$, respectively. \\
       & \hla{Primary} &  Since customers care about fairness and infer subproportionally , \hla{the price elasticity of demand} \\
        & \hlb{Ordered} & is \$\hlc{E} = \textbackslash e+(\textbackslash e-1) \textbackslash g \textbackslash f(M\^p(P))\$ .\\
       \hline
        \rota{6}{Relations} & Direct &  The \hla{dataset} with data-label pairs \$\hlc{D} =\{(x\_i, y\_i)\}\$\\
        & & Finally, the monopoly has a constant \hla{marginal cost} \$\hlc{C} > 0\$ \\
        \cline{2-3}
        & Coref-Description &  The dataset with \hla{data-label pairs} \$D=\{(x\_i, y\_i)\} \textbackslash cdots \$ . \hla{The few available labeled data} are called support set \$ \textbackslash mathcal\{S\}=\{(x\_i,y\_i)\}\$ \\
        \cline{2-3}
        & Coref-Symbol & It chooses price \$\hlc{P}\$ and output \$Y\$ to maximize profits \$(\hlc{P} - C) \textbackslash cdot Y\$ subject to customers' demand\\
        \cline{2-3}
        & Count &  $\cdots$ where every class in the task has $\hlc{q}$ \hla{test cases}. \\
        \hline
    \end{tabular}
    \label{tab:examples}
\end{table}

\textbf{Symbol tagging}: A mathematical symbol can present an operand, an operator, an expression, or combination of these.
\begin{itemize}
    \item An atomic symbol in PDF format has to be a character, that means, if there is ``$\backslash$hat\{Y\}'' in the document, neither ``Y'' nor ``$\backslash$hat'' is considered an atomic symbol, instead ``$\backslash$hat\{Y\}'' is a symbol.
    \item A complex symbol is a combination of multiple symbols and brackets, for example: ``P(x)'', ``Wx''
    \item An annotated symbol has to be a complete symbol e.g. ``P(x)'' is good, ``P(x'' is not because of lacking the closing parenthesis.
    \item A complex formula can be segmented into atomic symbols, we will annotate at all levels of the complex symbol as long as there are appropriate descriptions available.
\end{itemize}

\textbf{Relation annotation}: 

\begin{itemize} 
    \item Every annotated symbol/description has to have at least one relation linking to its description/symbol. 
    \item If there are multiple mentions of a single symbol/description, use coreference relation to link them. A direct relation or a count relation is used to link the closet pair of symbol and description.
\end{itemize}

\end{document}